%% file: neology_scil.tex
\documentclass[11pt,a4paper]{article}
\usepackage{authblk}
\usepackage[hyperref]{naaclhlt2019}
\usepackage{times}
\usepackage{latexsym}

\usepackage{url}
\usepackage{graphicx}
\usepackage{subcaption}
\usepackage{sidecap}
\usepackage{amsfonts}
\usepackage{amsmath}
\usepackage{booktabs}
\usepackage{multirow}
\usepackage{mathtools}

\aclfinalcopy % Uncomment this line for the final submission

\newcommand\modern{\textsc{modern}}
\newcommand\hist{\textsc{historical}}
\newcommand\form{\emph}

\newcommand\redtext{\textcolor{red}}
\renewcommand\redtext[1]{#1}

\newcommand{\Sref}[1]{\S\ref{#1}}

\newcommand{\Fref}[1]{Figure~\ref{#1}}

\newcommand{\Tref}[1]{Table~\ref{#1}}

\newcommand{\ignore}[1]{}

\title{Where New Words Are Born: Distributional Semantic Analysis of Neologisms and Their Semantic Neighborhoods}

\author[1]{Maria Ryskina}
\author[2]{Ella Rabinovich}
\author[3]{Taylor Berg-Kirkpatrick}
\author[1]{\protect\\ \bf{David R. Mortensen}} 
\author[1]{Yulia Tsvetkov}
\affil[1]{Language Technologies Institute, Carnegie Mellon University, \protect\\
{\normalsize\tt \{mryskina,dmortens,ytsvetko\}@cs.cmu.edu}}
\affil[2]{Department of Computer Science, University of Toronto, {\tt ella@cs.toronto.edu}}
\affil[3]{Computer Science and Engineering, University of California, San Diego, {\tt tberg@eng.ucsd.edu}}

\date{}

\begin{document}
\maketitle
\begin{abstract}
  We perform statistical analysis of the phenomenon of \emph{neology}, the process by which new words emerge in a language, using large diachronic corpora of English. We investigate the importance of two factors, semantic sparsity and frequency growth rates of semantic neighbors, formalized in the distributional semantics paradigm. We show that both factors are predictive of word emergence although we find more support for the latter hypothesis. Besides presenting a new linguistic application of distributional semantics, this study tackles the linguistic question of the role of language-internal factors (in our case, sparsity) in language change motivated by language-external factors (reflected in frequency growth).\footnote{The code and word lists are available at \url{https://github.com/ryskina/neology}}
\end{abstract}

\defcitealias{oed}{OED, 2018}	

\section{Introduction}

Natural languages are constantly changing as the context of their users changes \cite{aitchison2001language}. Perhaps the most obvious type of change is the introduction of new lexical items, or \emph{neologisms} (a process called ``neology''). Neologisms have various sources. They are occassionally coined out of whole cloth (\emph{grok}). More frequently, they are loanwords from another language (\emph{tahini}), derived words (\emph{unfriend}), or existing words that have acquired new senses (as when \emph{web} came to mean `World Wide Web' and then `the Internet').
While neology has long been of interest to linguists (\Sref{sec:background}), there have been relatively few attempts to study it as a global, systemic phenomenon. Computational modeling and analysis of neology is the focus of our work.

What are the factors that predict neology? Certainly, social context plays a role. Close interaction between two cultures, for example, may result in increased borrowing \cite{appel2006language}. We hypothesize, though, that there are other factors involved---factors that can be modeled more directly. These factors can be understood in terms of \textbf{supply} and \textbf{demand}.
 
\newcite{breal1904essai} introduced the idea that %words exist in a semantic space. His work implies that, over time and other things being equal, 
the distribution of words in semantic space tends towards uniformity. 
%This is most explicit in his law of differentiation, which states that near synonyms move apart in semantic space, but his framework leaves open space for other repairs. For example, 
This framework predicts that new words would emerge where they would repair uniformity---where there was a space not occupied by a word. This could be viewed as supply-driven neology. Next, demand plays a role as well as supply \cite{campbell2013historical}: %. In particular, we posit that 
new words emerge in ``stylish'' neighborhoods, corresponding to domains of discourse that are increasing in importance (reflected by the increasing frequency of the words in those neighborhoods).

%\yt{paragraph 4: methodology, connecting it with Bre\'al idea + integrating Taylor's first comment 
%-word embeddings provide us a nice way to operationalize breal's hypothesis and locate embeddings and their semantic neighborhood and computationally analyze the \textbf{supply} part
%forward reference to hypothesis 1 
%how we implement it? 
%we look at sparsity }
We operationalize these ideas using distributional semantics \cite{lenci2018distributional}. %, a paradigm in which words are represented as embedding vectors reflecting their co-occurrence statistics 
%Word embeddings provide a convenient way to locate words in the semantic space and analyze their semantic neighborhoods. 
To formalize the hypothesis of supply-driven neology for computational analysis, we measure \textbf{sparsity of areas in the word embedding space} where neologisms would later emerge. %; this is our first hypothesis. %, defined more formally in~\Sref{sec:hypotheses}.
%\yt{paragraph 5: moving to hypothesis 2+its implementation
%david's housing metaphor as a motivation
%we computationally analyze the \textbf{demand} part
%forward reference to hypothesis 2
%how we implement it \\}
%But just as there are factors that predict where houses are built other than the availability of land, we propose that there are factors that predict where new words emerge other than the availability of semantic space. 
The demand-driven view of neology motivates our second hypothesis: \textbf{neighborhoods in the embedding space containing words rapidly growing in frequency} are more likely to produce neologisms. Both hypotheses are defined more formally in \Sref{sec:hypotheses}. 

%\yt{paragraph 6: 
%experiments (\Sref{sec:method})
%results (\Sref{sec:results})
%contributions 
%-- into NLP: how we extend existing research on semantic change
%-- inlo linguistics: ask David to phrase it}
Having formalized our hypotheses in terms of word embeddings, we test them by comparing the distributions of the corresponding metrics for a set of automatically identified neologisms and a control set. Methodology of the word selection and hypothesis testing is detailed in \Sref{sec:method}. We discuss the results in \Sref{sec:results}, demonstrating evidence for both hypotheses, although the demand-driven hypothesis has more significant support.

\vspace{-0.1cm}
\section{Background}
\label{sec:background}
\noindent\textbf{Neology} 
Specific sources of neologisms have been studied: lexical borrowing \cite{taylor2014lexicalborrowing,daulton2012lexical}, morphological derivation \cite{lieber2017derivational}, blends or portmanteaus \cite{cook2012using,renner2012cross}, clippings, acronyms,  analogical coinages, and arbitrary coinages, but these studies have tended to look at neologisms atomistically, or to explicate the social conditions under which a new word entered a language rather than looking at neologisms in systemic context.

To address this deficit, we look back to the seminal work of Michel Br\'eal, who introduced the idea that words exist in a semantic space. His work implies that, other things being equal, the semantic distribution of words tends towards uniformity~\cite{breal1904essai}. This is most explicit in his law of differentiation, which states that near synonyms move apart in semantic space, but has other implications as well. For example, this principle predicts that new words are more likely to emerge where they would increase uniformity. This could be viewed as supply-driven neology---new words appear to fill gaps in semantic space (to express concepts that are not currently lexicalized). %This factor is operationalized as the sparcity of the neighborhood in which a neologism emerges.

In linguistic literature neology is often associated with new concepts or domains of increasing importance \cite{campbell2013historical}. Just as there are factors that predict where houses are built other than the availability of land, there are factors that predict where new words emerge other than the availability of semantic space. Demand, we hypothesize, plays a role as well as supply. %In particular, we posit that new words emerge in ``stylish'' neighborhoods, corresponding to domains of discourse that are increasing in importance (reflected by the increasing frequency of the words in those neighborhoods). In both linguistic literature and common parlance, 
 %\textbf{Demand-driven neology} is our formalization of this widely hypothesized (but empirically unvalidated) phenomenon.

%\section{Related Work}

Most existing computational research on the mechanisms of neology focuses on discovering sociolinguistic factors that predict acceptance of emerging words into the mainstream language and growth of their usage, typically in online social communities~\cite{del2018road}.
The sociolinguistic factors can include geography~\cite{eisenstein2017identifying}, user demographics~\cite{eisenstein2012mapping, eisenstein2014diffusion}, diversity of linguistic contexts~\cite{stewart2018making} or word form~\cite{kershaw2016towards}.
To the best of our knowledge, there is no prior work focused on discovering factors predictive of the emergence of new words rather than modeling their lifecycle. %We also rely entirely on language-internal factors, modeling language-external processes indirectly through their reflection in the language.
We model language-external processes indirectly through their reflection in language, thereby capturing phenomena evident of our hypotheses through linguistic analysis.

%To  the  best  of  our  knowledge,  there  is no  prior  work  that  would  abstract  away  from the  information  contained  in  the  distribution  of a  specific neologism  and focus  on analyzing  the emergence of new words solely from evolution of existing word forms.
%\note{the last part of the sentence (from evolution...) is not completely clear to me}  
%\tbk{Is the key here that we focus on language-internal factors, while others look at external factors?} 
%\drm{I think that the language-internal/language-external contrast is good but we have to be careful---H2 could be seen as a reflection of language-external factors if we don't frame it right.}

\noindent\textbf{Distributional semantics and language change} Word embeddings have been successfully used for different applications of the diachronic analysis of language~\cite{tahmasebi2018survey}. The closest task to ours is analyzing meaning shift (tracking changes in word sense or emergence of new senses) by comparing word embedding spaces across time periods~\cite{kulkarni2015statistically, xu2015computational, hamilton2016diachronic, kutuzov2018diachronic}. Typically, embeddings are learned for discrete time periods and then aligned \citep[but see][]{bamler2017dynamic}.
%\citet{bamler2017dynamic} extend the approach by proposing a dynamic embedding model that learns continuous embedding trajectories over time.
There has also been work on revising the existing methodology, specifically accounting for frequency effects in embeddings when modeling semantic shift~\cite{dubossarsky2017outta}. 
 
Other related questions where distributional semantics proved useful were exploring the evolution of bias~\cite{garg2018word} and the degradation of age- and gender-predictive language models~\cite{jaidka2018diachronic}. 
%\comment{\tbk{Maybe we're also the first to use word embeddings as a way of quantifying semantic sparsity?}}
%\\

% { \color{red}
% Other possibly relevant cites:
% \cite{frermann2016bayesian} - Bayesian model of diachronic meaning change as continuous process (not using embeddings)
% }

%\clearpage
% charts were here (Figure 1)

\section{Hypotheses \label{sec:hypotheses}}
This section outlines the two hypotheses we introduced earlier from the linguistic perspective, formalized in terms of distributional semantics. 

%\tbk{Main idea: First, state our hypotheses in linguistic terms, citing prior linguistics work that introduced these ideas. Finally, translate these linguistic hypotheses into what we expect to see in embedding spaces. We do some of both already, but could add citations and more background on each linguistic hypothesis first.}

\paragraph{Hypothesis 1} \textit{Neologisms are more likely to emerge in sparser areas of the semantic space.} This corresponds to the supply-driven neology hypothesis: we assume that areas of the space that contain fewer semantically related words are likely to give birth to new ones so as to fill in the `semantic gaps'. Word embeddings give us a natural way of formalizing this: since semantically related words have been shown to populate the same regions in embeddings spaces, %we can translate  semantic sparsity into geometric sparsity, measuring density of the semantic neighborhood of a word as number of word vectors within a certain distance of the word's embedding.
we can approximate semantic sparsity (or density) of a word's neighborhood as the number of word vectors within a certain distance of its embedding.

%To formalize this hypothesis, we use distributional semantics, a paradigm in which words are represented as embedding vectors reflecting their co-occurrence statistics \cite{bullinaria2007extracting, turney2010frequency}. We assume that sparsity of the neighborhood in the embedding space can be representative of lexical need in this semantic area.
%\tbk{Can we say a bit more about what sparisty means in the linguistic sense? (i.e. sparse areas of the semantic space are those where fewer words have related meanings) Then we can say embedding spaces give us a nice way to formalize this: semantic sparsity can be translated to geometric sparsity.}

\paragraph{Hypothesis 2} \textit{Neologisms are more likely to emerge in semantic neighborhoods of growing popularity.} Here we formalize our demand-driven view of neology, which assumes that growing frequency of words in a semantic area is a reflection of its growing importance in discourse, and that the latter is in turn correlated with emergence of neologisms in that area.
In terms of word embeddings, we again consider nearest word vectors as the word's semantic neighbors and quantify the rate at which their frequencies grow over decades (formally defined in \Sref{sec:control}).

\section{Methodology \label{sec:method}}
Our analysis is based on comparing embedding space neighborhoods of neologism word vectors and neighborhoods of embeddings of words from an alternative set. Automatic selection of neologisms is described in \Sref{sec:select}, and in \Sref{sec:control} we detail the factors we control for when selecting the alternative set. In \Sref{sec:data} we describe the datasets used in our experiments. Our data is split into two large corpora, \hist{} and \modern{}; we additionally require the \hist{} corpus to be split into smaller time periods so that we can estimate word frequency change rate. Embedding models are trained on each of the two corpora, as described in \Sref{sec:embedding}. We compare the neighborhoods in the \hist{} embedding space, but due to the nature of our neologism selection process, many neologisms might not exist in the \hist{} vocabulary. To locate their neighborhoods, we adapt an approach from prior work in diachronic analysis with word embeddings: we learn an orthogonal projection between \hist{} and \modern{} embeddings to align the two spaces in order to make them comparable~\citep[see][]{hamilton2016diachronic}, and use projected vectors to represent neologisms in the \hist{} space. Finally, \Sref{sec:setup} describes the details of hypothesis testing: statistics we choose to quantify our two hypotheses and how their distributions are compared.

\subsection{Datasets \label{sec:data}}
We use the Corpus of Historical American English (COHA, \citealp{davies2002coha}) and the Corpus of Contemporary American English (COCA, \citealp{davies2008coca}), large diachronic corpora balanced by genre to reflect the variety in word usage. COHA data is split into decades; we group COHA documents from 18 decades (1800-1989) to represent the \hist{} English collection and use full COCA 1990-2012 corpus as \modern{}.

The obtained \hist{} split contains 405M tokens of 2M types, and \modern{} contains 547M tokens of 3M types.\footnote{Statistics accompanying the corpora state that entire COHA dataset contains 385M words, and COCA contains 440M words; we assume the discrepancy is explained by tokenization differences.}

\subsection{Neologism selection \label{sec:select}}
\redtext{We rely on a usage-based approach to extract the set of neologisms for our analysis, choosing the words based on their patterns of occurrence in our datasets. It can be seen as an approximation to selecting words based on their earliest recorded use dates, as these dates are also determined based on the words' usage in historical corpora. This analogy is supported by the qualitative analysis of the obtained set of neologisms, as discussed in \Sref{sec:discussion}.}

We limit our analysis to nouns, an open-class lexical category.
% I was not able to track down a citation saying that neologisms are most often nouns, though I know such claims have been made in the literature. We need to finess this somehow.
We identify nouns in our corpora using a part-of-speech dictionary, collected from a POS-tagged corpus of English Wikipedia data~\citep[Wikicorpus,][]{reese2010wikicorpus},
and select words that are most frequently tagged as `NN'.

We additionally filter candidate neologisms to exclude words that occur more frequently in capitalized than lowercased form; this heuristic helps us remove proper nouns missed by the POS tagger.

We select a set of neologisms by picking words that are substantially more frequent in the \modern{} corpus than in the \hist{} one. It is important to note that while we use the term ``neologism,'' implying a word at the early stages of emergence, with this method we select words that have entered mainstream vocabulary in \modern{} time but might have been coined prior to that. We consider a word $w$ to be a neologism if its ratio $f_m(w) / f_h(w)$ is greater than a certain threshold; here $f_m(\cdot)$ and $f_h(\cdot)$ denote word frequencies (normalized counts) in \modern{} and \hist{} data respectively. Empirically we set the frequency ratio threshold equal to 20. 
%vSample neologisms along with the qualitative analysis of the obtained set can be found in \Sref{sec:discussion}.

We rank words satisfying these criteria by their frequency in the \modern{} corpus and select the first 1000 words to be our neologism set; this is to ensure that we only analyze words that subsequently become mainstream and not misspellings or other artifacts of the data.
%\footnote{Obtained list of neologisms is included in supplementary material.} 

\subsection{Embeddings \label{sec:embedding}}
Our hypothesis testing process involves inspecting semantic neighborhoods of neologisms in the \hist{} embedding space. However, many neologisms are very infrequent or nonexistent in the \hist{} data, so we approximate their vectors in the \hist{} space by projecting their \modern{} embeddings into the same coordinate axes.

We learn Word2Vec Skip-Gram embeddings\footnote{Hyperparameters: vector dimension 300, window size 5, minimum count 5.}~\cite{mikolov2013distributed} of the two corpora and use orthogonal Procrustes to learn the aligning transformation:
$$
\mathbf{R} = \arg \min_{\mathbf{\Omega}} \|\mathbf{\Omega W}^{(m)} - \mathbf{W}^{(h)} \|, 
$$
where $\mathbf{W}^{(h)}, \mathbf{W}^{(m)} \in \mathbb{R}^{|V| \times d}$ are the word embedding matrices learned on the \hist{} and \modern{} corpora respectively, restricted to the intersection of the vocabularies of the two corpora (i.e. every word embedding present in both spaces is used as an anchor). To project \modern{} word embeddings into the \hist{} space, we multiply them by the obtained rotation matrix $\mathbf{R}$.

% neighborhood figure was here
\begin{figure*}[!t]
    \centering
    \begin{subfigure}[b]{0.48\textwidth}
        \includegraphics[width=\textwidth]{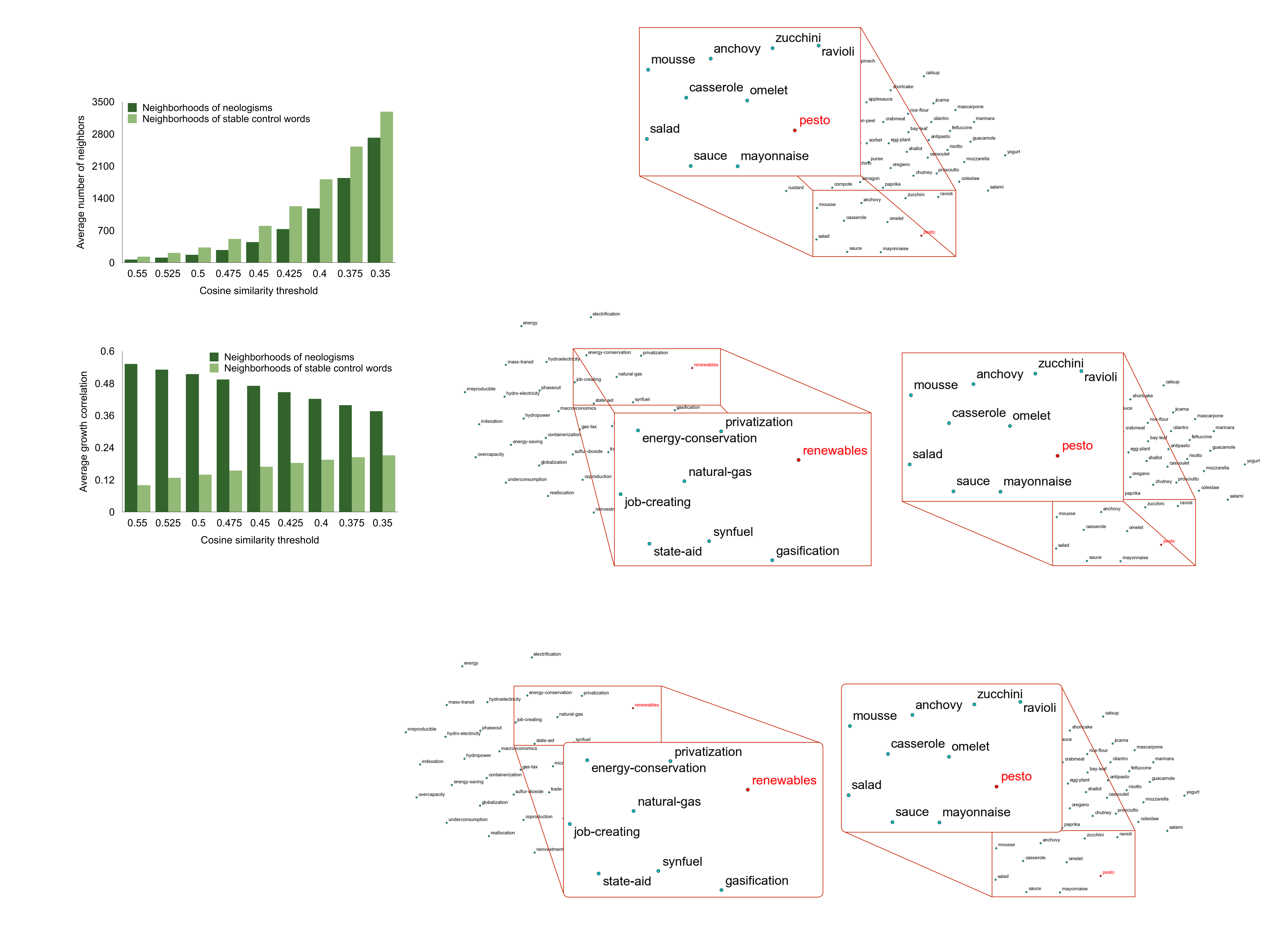}
        \caption{\label{fig:renewables} Semantic neighborhood of the word \form{renewables}.}
    \end{subfigure}
    \hspace{0.3cm}
    \begin{subfigure}[b]{0.45\textwidth}
        \includegraphics[width=\textwidth]{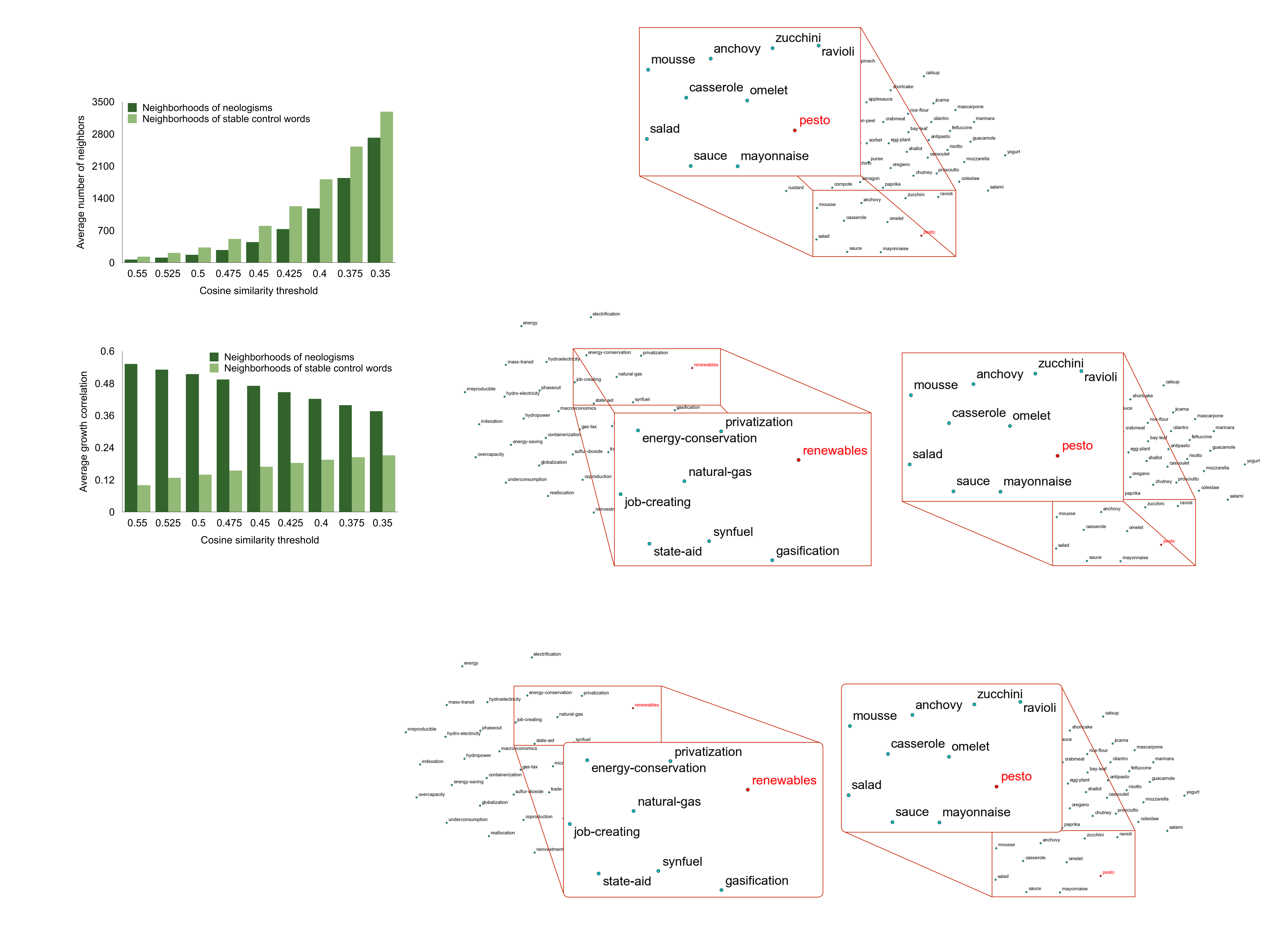}
        \caption{\label{fig:pesto} Semantic neighborhood of the word \form{pesto}.}
    \end{subfigure}
    \caption{\label{fig:neighbors} Neighborhoods of projected \modern{} embeddings of two neologisms (shown in red), \form{renewables} and \form{pesto}, in the \hist{} embedding space, visualized using t-SNE~\cite{maaten2008visualizing}. Figure~\ref{fig:renewables} shows an example of a neighborhood exhibiting frequency growth: words like \form{synfuel} or \form{privatization} have been used more towards the end of the \hist{} period. The neighborhood also includes \form{natural-gas} that can be seen as representing a concept to be replaced by \form{renewables}. The word \form{pesto} (Figure~\ref{fig:pesto}) is projected into a neighborhood of other food-related words, most of which are also loanwords, several from the same language; it also has its hypernym \form{sauce} as one of its neighbors.}
\end{figure*}

\subsection{Control set selection}\label{sec:control}
To test our hypotheses, we collect an alternative set of words and analyze how certain statistical properties of their neighbors differ from those of neighbors of neologisms. At this stage it is important to control for non-semantic confounding factors that might affect the word distribution in the semantic space. One such factor is word frequency: it has been shown that embeddings of words of similar frequency tend to be closer in the embedding space~\cite{schnabel2015evaluation, faruqui2016problems}, which results in very dense clusters, or hubs, of words with high cosine similarity~\cite{radovanovic2010hubs, dinu2014improving}.
We choose to also restrict our control set to only include words that did not substantially grow or decline in frequency over the \hist{} period in order to prevent selecting counterparts that only share similar frequency in the \modern{} sub-corpus (e.g., due to recent topical relevance), but exhibit significant fluctuation prior to that period. In particular, we refrain from selecting words that emerged in language right before our \hist{}-\modern{} split.

We create the alternative set by pairing each neologism with a non-neologism counterpart that exhibits a stable frequency pattern, while controlling for word frequency and word length in characters. Length is chosen as an easily accessible correlate to other factors for which one should control, such as morphological complexity, concreteness, and nativeness.
We perform the pairing only to ensure that the distribution of those properties across the two sets is comparable, but once the selection process is complete we treat control words as a set rather than considering them in pairs with neologisms.

Following~\citet{stewart2018making}, we formalize frequency growth rate as the Spearman correlation coefficient between timesteps $\{1, \ldots, T\}$ and frequency series $f_{(1:T)}(w)$ of word $w$. In our setup, timesteps $\{1, \ldots, 18\}$ enumerate decades from 1810s to 1980s, and $f_t(\cdot)$ denote word frequencies in the corresponding $t$-th decade of the \hist{} data.
%\tbk{I agree that we need to motivate stability somehow -- maybe we can just say that in our most restricted experiments, we want to check how neologisms differ from words that are neither increasing or decreasing in freq. Then, when we relax the stability constraint, we check how neologisms differ from non-neologisms more broadly.}

Formally, for each neologism $w_n$ we select a counterpart $w_c$ satisfying the following constraints:
\begin{itemize}
    \item Frequencies of the two words in the corresponding corpora are comparable: $f_m(w_n) / f_h(w_c) \in (1-\delta, 1+\delta)$, where $\delta$ was set to 0.25;
    \item The length of the two words is identical up to 2 characters;
    \item The Spearman correlation coefficient $r_s$ between decades $\{1, \ldots, 18 \}$ and the control word frequency series $f_{(1:18)}(w_c)$ is small: $|r_s\left(\{1:18\}, f_{(1:18)}(w_c)\right)| \leq 0.1$
    %\tbk{No term in the series exceeds 0.1?}
    %The series of frequencies of the control word across decades and the \redtext{series of decade indices} do not have a strong correlation: $|r(F, D)| < 0.1$, where $F =\{f_t(w_c)\}_{t=1}^{18}$ and $D = \{1, \ldots, 18\}$ 
    %\note{please have a look at how they describe a similar procedure in \url{https://arxiv.org/pdf/1709.00345.pdf} section 3.1; also `18' should be explained}
\end{itemize}
These words, which we will refer to as \emph{stable}, make up our default and most restricted control set. We will also compare neologisms to a \emph{relaxed} control set, omitting the stability constraint on the frequency change rate but still controlling for length and overall frequency, to see how neologisms differ from non-neologisms in a broader perspective.
%\tbk{Should we introduce, at this point, the fact that we will use two different kinds of controls? One set with stability, the other without?}
%\drm{I agree with Taylor's suggestion; I think it will make it less confusion later on.}

\subsection{Experimental setup \label{sec:setup}}
We evaluate our hypotheses by inspecting neighborhoods of neologisms and their stable control counterparts in the \hist{} embedding space, viewing them as proxy for neighborhoods in the underlying semantic space. Since many neologisms are very infrequent or nonexistent in the \hist{} data, we approximate their vectors in the \hist{} space with their \modern{} embeddings projected using the transformation described in \Sref{sec:embedding}. The neighborhood of a word $w$ is defined as the set of \hist{} words for which cosine similarity between their \hist{} embeddings and $v_w$ exceeds the given threshold $\tau$; $v_w$ denotes a projected \modern{} embedding if $w$ is a neologism or a \hist{} embedding if it is a control word.\footnote{Cosine similarity is chosen as our distance metric since it is traditionally used for word similarity tasks in distributional semantics~\cite{lenci2018distributional}. \redtext{We have also observed the same results when repeating the experiments with the Euclidean distance metric.}}

The two factors we need to formalize are semantic sparsity of the neighborhoods and increase of popularity of the topic that the neighborhood represents. We use sparsity in the embedding space as a proxy for semantic sparsity and approximate growth of interest in a topic with frequency growth of words belonging to it (i.e. embedded into the corresponding neighborhood). For the neighborhood of each word $w$, we compute the following statistics, corresponding to our two hypotheses:
%\tbk{Can we say a little more about these statistics? For example, why do we use cosine distance?}
\begin{enumerate}
    \item \emph{Density of a neighborhood} $d(w, \tau)$: number of words that fall into this neighborhood $d(w, \tau) = |\{u: \text{cosine}(v_w, v_u) \geq \tau\}|$
    \item \emph{Average frequency growth rate of a neighborhood} $r(w, \tau)$: as defined in the previous subsection, we compute the Spearman correlation coefficient between timesteps and frequency series for each word in the neighborhood and take their mean: 
    \begin{align*}
        r(w, \tau) = & \frac{1}{d(w, \tau)} \times \\ 
        & \times \smashoperator[lr]{\sum\limits_{u: \text{cosine}(v_w, v_u) \geq \tau}} r_s \left( \{1:18\}, f_{(1:18)}(u) \right)
    \end{align*}
\end{enumerate}

In our tests, we compare the values of those metrics for neighborhoods of neologisms and neighborhoods of control words and estimate the significance of each of the two factors for a range of neighborhood sizes defined by the threshold $\tau$. We test whether means of the distributions of those statistics for the neologism and the control set differ and whether each of the two is significant for classifying words into neologisms and controls.

%\tbk{Here, I think we could add a paragraph about what kinds statistical tests we are going to perform to attempt to check our hypothesis. So far, all we say is that we compute these two statistics, we don't yet say how we're going to use them more specifically. You could give a high-level overview of the types of hypothesis tests we're going to run.}
 
As mentioned in \Sref{sec:select}, our vocabulary is restricted to nouns, and we only consider vocabulary noun neighbors when evaluating the statistics.\footnote{Here we refer to the vocabulary of words participating in our analysis, not the embedding model vocabulary; embeddings are trained on the entire corpora.} Since we project all neologism word vectors from \modern{} to \hist{} embedding space, for neologisms occurring in the \hist{} corpus we might find a \hist{} vector of the neologism itself among the neighbors of its projection; we exclude such neighbors from our analysis. We cap the number of nearest neighbors to consider at 5,000, to avoid estimating statistics on overly large sets of possibly less relevant neighbors.

\begin{figure*}[!t]
    \centering
    \begin{subfigure}[b]{0.48\textwidth}
        \includegraphics[width=\textwidth]{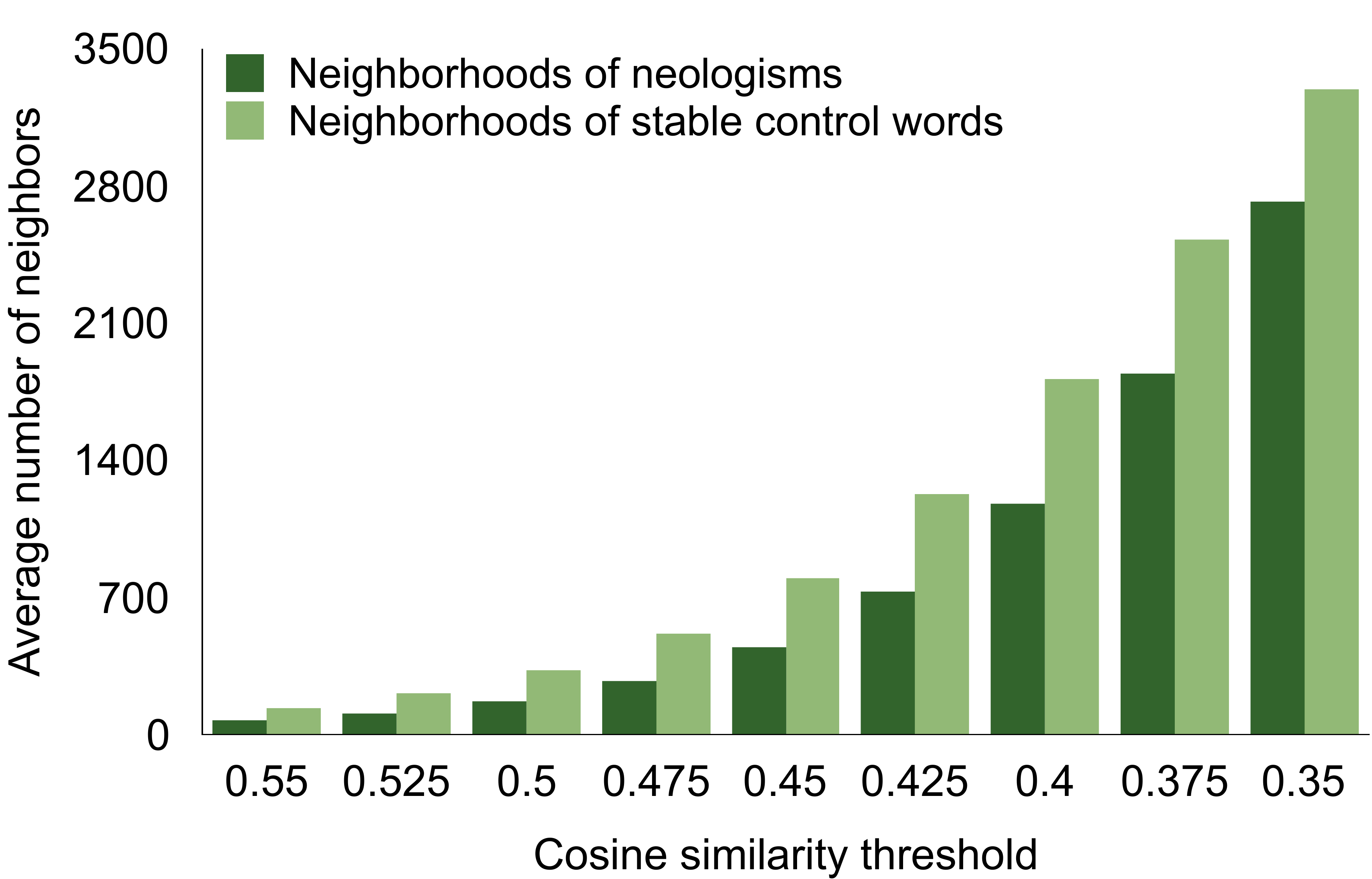}
        \caption{\label{fig:density} Average \hist{} word vector density in the neighborhoods of neologisms and stable control set words.\\}
    \end{subfigure}
    \hspace{.3cm}
    \begin{subfigure}[b]{0.48\textwidth}
        \includegraphics[width=\textwidth]{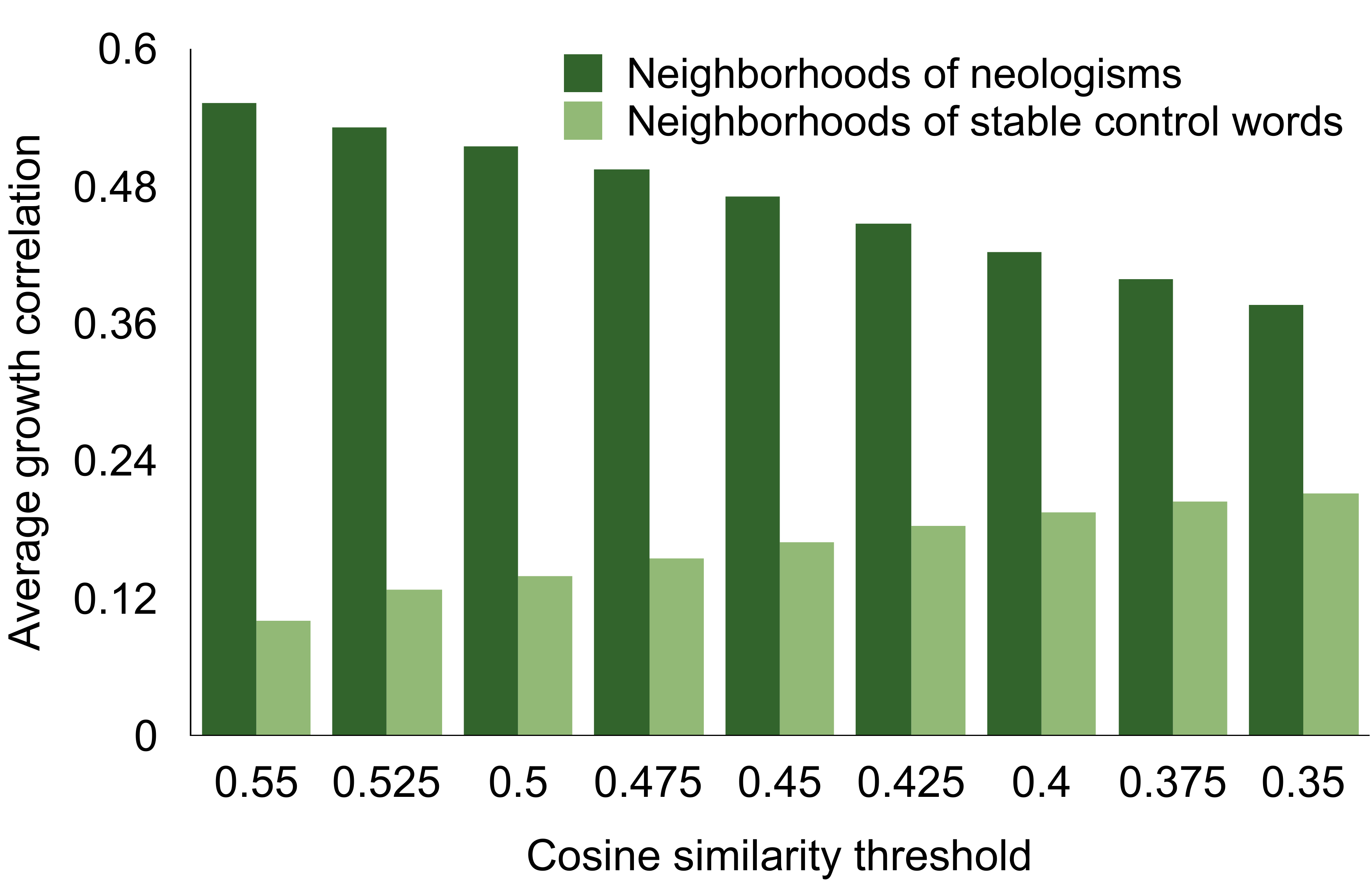}
        \caption{\label{fig:growth} Average frequency growth rate of \textsc{historical} word vectors in the neighborhoods of neologisms and stable control set words.}
    \end{subfigure}
    \caption{\label{fig:charts} Number of \hist{} word vectors within a certain cosine distance of a word and average growth rate of frequency (represented by Spearman correlation coefficient) of those \hist{} words, averaged across neologism (darker) and stable control word (lighter) sets. Projected neologism vectors appear in lower-density neighborhoods compared to control words, and neighbors of neologisms exhibit a stronger growth trend than those of the control words, especially in smaller neighborhoods.}
\end{figure*}

\section{Results \label{sec:results}}
Following the experimental setup described in \Sref{sec:setup}, we estimate the contribution of each of the hypothesized factors employing strictly constrained and relaxed control sets. We start by analyzing how the distributions of those statistics differ for neologisms and stable controls, both by comparing their sample means and by more rigorous statistical testing. We also evaluate the significance of the factors using generalized linear models for both stable and relaxed control sets.

\subsection{Comparison to stable control set}
First, we test our hypotheses on 720 neologism-stable control word pairs (not all words are paired in the stable control setting due to its restrictiveness).
%, computing densities and average growth rates for a range of neighborhood sizes, defined by the cosine similarity threshold.

%\comment{\note{I know we ended up with ~720, would we have the whole 1000 if releasing the 0.25 a bit? say $\alpha$ is 0.35? or perhaps let's go with the top-750 (and not 1000) at the first place?} \redtext{\textit{I will try running that tomorrow. I don't think restricting to 750 most frequent will give us a ful set of pairs, it's more of a length issue that a frequency one.}}.}

\Fref{fig:charts} demonstrates the values of density and frequency growth rate for a range of neighborhood sizes, averaged over neologism and control sets. Both results conform with our hypotheses: \Fref{fig:density} shows that on average the projected neologism has fewer neighbors than its stable counterpart, especially for larger neighborhoods, and \Fref{fig:growth} shows that, on average, frequencies of neighbors of a projected neologism grow at a faster rate than those of a counterpart. Interestingly, we find that neighbors of stable controls still tend to exhibit small positive growth rate. We attribute it to the general pattern that we observed: about 70\% of words in our vocabulary have positive frequency growth rate. 
We believe this might be explained by the imbalance in the amount of data between decades (e.g. 1980s sub-corpus has 20 times more tokens than 1810s): some words might not occur until later in the corpus because of the relative sparsity of data in the early decades.

As we can see from \Fref{fig:density}, neighborhoods of larger sizes (corresponding to lower values of the threshold) may contain thousands of words, so the statistics obtained from those neighborhoods might be less relevant; we might only want to consider the immediate neighborhoods, as those words are more likely to be semantically related to the central word. It is notable that the difference in the growth trends of the neighbors is substantially more prominent for smaller neighborhoods (\Fref{fig:growth}): average correlation coefficient of immediate neighbors of stable words also falls into stable range as we defined it, while immediate neighbors of neologisms exhibit rapid growth.

%We test the significance of our results for both hypotheses (Figures~\ref{fig:density} and~\ref{fig:growth}) by applying the Wilcoxon signed-rank test on two series of values: (1) neighborhood densities of neologisms and those of their control set counterparts, and (2) neighborhood frequency growth rates
%values representing neighborhood growth of neologisms and those of their control set pairs. While the differences in growth values are significant at all radii (\Fref{fig:growth}) at the level of $p<0.001$, only the two largest neighborhoods show significant differences in density (Figure ~\ref{fig:density}).

\subsection{Statistical significance}

\input{glm_table.tex}

To estimate the significance and relative contribution of the two factors, we fit a generalized linear model (GLM) with logistic link function to the corresponding features of neologism and control word neighborhoods:\footnote{We use the implementation provided in the MATLAB Statistics and Machine Learning Toolbox.}
\begin{multline*}
    y(w) \sim (1 + \exp(-\beta_0^{(\tau)} - \\ - \beta_d^{(\tau)} \cdot d(w, \tau)  - \beta_r^{(\tau)} \cdot r(w, \tau)))^{-1}
\end{multline*}
where $y$ is a Bernoulli variable indicating whether the word $w$ belongs to the neologism set (1) or the control set (0), and $\tau$ is the cosine similarity threshold defining the neighborhood size.

\Tref{tab:glm} shows how the coefficients and $p$-values for the two statistics change with the neighborhood size. We found that when comparing with the stable control set, average frequency growth rate of the neighborhood was significant for all sizes, but neighborhood density was significant at level $p<0.01$ only for the largest ones.\footnote{Applying Wilcoxon signed-rank test to the series of neighborhood density and frequency growth values for neologism and stable control sets showed the same results.} We attribute this to the effect discussed in the previous section: difference in average frequency growth rate between neighbors of neologisms and stable words shrinks as we include more remote neighbors (\Fref{fig:growth}), so for large neighborhoods frequency growth rate by itself is no longer predictive enough.

We also evaluate the significance of features for the relaxed control set without the stability constraint on 1000 neologism-control pairs.
%In this setting, both quantities are significant at level $p<0.01$ for any neighborhood size. 
We have repeated the experiment with 5 different randomly sampled relaxed control sets (results for one showed in~\Tref{tab:glm}). For medium-sized neighborhoods ($0.4 \leq \tau \leq 0.5$) density variable is always significant at $p < 0.01$, but densities of largest and smallest neighborhoods were rejected in several runs. With more variance in the control set, differences in neighborhood frequency growth rate between neologisms and controls are less prominent than in the stable setting, so density plays a more important role in prediction.\footnote{\redtext{Detailed results of the regression analysis and collinearity tests can be found in the repository. No evidence of collinearity was found in any of the experiments.}}
%(but significant at $p < 0.05$).
%Densities of largest and smallest neighborhoods were significant at $p < 0.01$ only for two runs, and four times for $p < 0.05$.
%At $p<0.01$, densities of smallest and largest neighborhoods have been found insignificant 3 out of 5 times but for medium-sized neighborhoods ($0.4 \leq \tau \leq 0.5$) density is always significant. 
%At $p<0.05$ only the density of the largest neighborhood has been rejected once.  

Growth feature weights $\beta_r^{(\tau)}$ are always positive and density feature weights $\beta_d^{(\tau)}$ are negative in the relaxed setting (where density is significant). This matches our intuition that neighborhood frequency growth and sparsity are predictive of neology.
%\redtext{We do not interpret positive weights on density in the stable setting as contradictory to our hypothesis, since the feature did not prove significant in this setting.}
%\todo{Explain that positive weight on density doesn't matter when it's not significant.}

Comparing sample means of density and growth rates between neologisms and each of the 5 randomly selected relaxed control sets (as we did for stable controls in \Fref{fig:charts}) demonstrated that neologisms still appear in sparser neighborhoods than the controlled counterparts. The difference in frequency growth rate between the neologism and control word neighborhoods is also observed for all control sets (although it varies noticeably between sets), but it no longer exhibits an inverse correlation with neighborhood size.

\section{Discussion \label{sec:discussion}}

We have demonstrated that our two hypotheses hold for the set of words we automatically selected to represent neologisms. To establish validity of our results, we qualitatively examine the obtained word list to see if the words are in fact recent additions to the language. We randomly sample 100 words out of the 1000 selected neologisms and look up their earliest recorded use in the Oxford English Dictionary Online~\citepalias{oed}. Of those 100 words, eight are not defined in the dictionary: they only appear in quotations in other entries (\form{bycatch} (quotation from 1995), \form{twentysomething} (1997), \form{cross-sex} (1958), etc.) or do not occur at all (\form{all-mountain}, \form{interobserver}, \form{off-task}). Of the remaining 92 words, 78 have been first recorded after the year 1810 (i.e. since the beginning of the \hist{} timeframe), 44 have been first recorded in the twentieth century, and 21 words since 1950. However, some of the words dating back to before 19th century have only been recorded in their earlier, possibly obsolete sense: for example, while there is evidence of the word \form{software} being used in 18th century, this usage corresponds to its obsolete meaning of `textiles, fabrics', while the first recorded use in its currently dominant sense of `programs essential to the operation of a computer system' is dated 1958. To account for such semantic neologisms, we can count the first recorded use of the newest sense of the word; that gives us 82, 58 and 31 words appearing since 1810, 1900 and 1950 respectively.\footnote{For all words that have one or more senses marked as a noun, we only consider those senses. Out of the 92 listed words, only three do not have nominal senses, and for two more usage as a noun is marked to be rare.} This leads us to assume that most words selected for our analysis have indeed been neologisms sometime over the course of the \hist{} time. 

We would also like to note that the results of this examination may be skewed due to factors for which lexicography may not account: for example, many words identified as neologisms are compound nouns like \form{countertop} or \form{soundtrack} that have been written as two separate words or joined with a hyphen in earlier use. There is also considerable spelling variation in loanwords, e.g. \form{cuscusu}, \form{cooscoosoos}, \form{kesksoo} were used interchangeably before the form \form{couscous} was accepted as the standard spelling. Specific word forms might also have different life cycles: while the word \form{music} existed in Middle English, the plural form \form{musics} in a particular sense of `genres, styles of music' is much more recent. 

Qualitative examination of the neologism set reveals that new words tend to appear in the same topics; for example, many words in our set were related to food, technology, or medicine. This indirectly supports our second hypothesis: rapid change in these spheres makes it likely for related terms to substantially grow in frequency over a short period of time. One example of such a neighborhood is shown in \Fref{fig:renewables}: the neologism \form{renewables} appeared in a cluster of words related to energy sources --- a topic that has been more discussed recently. There is also some correlation between the topic and how new words are formed in it: most food neologisms are so-called cultural borrowings~\cite{weinreich2010languages}, when the name gets loaned from another culture together with the concept itself (e.g. \form{pesto}, \form{salsa}, \form{masala}), while many technology neologisms are compounds of existing English morphemes (e.g. \form{cyber+space}, \form{cell+phone}, \form{data+base}). 

We also consider nearest neighbors (\hist{} words with highest cosine similarity) of the neologisms to ensure that they are projected into the appropriate parts of the embedding space. Examples of nearest neighbors are shown in \Tref{tab:nns}. We saw different patterns of how the concept represented by the neologism relates to concepts represented by its neighbors. For example, some terms for new concepts appear next to related concepts they succeeded and possibly made obsolete: e.g. \form{email:letter}, \form{e-book:paperback}, \form{database:card-index}. Other neologisms emerge in clusters of related concepts they still equally coexist with: \form{hip-hop:jazz}, \form{hoodie:turtleneck}; most cultural borrowings fall under this type (see the neighborhood of \form{pesto} in \Fref{fig:pesto}). Both those patterns can be viewed as examples of a more general trend: one concept takes place of another related one, whether in terms of fully replacing it or just taking its place as the dominant form. %\todo{"X is the new Y"} 

Other interesting effects we observed include lexical replacement (a new word form replacing an old one without a change in meaning, e.g. \form{vibe:ambience}), tendency to abbreviate terms as they become mainstream (\form{biotech:biotechnology}, \form{chemo:chemotherapy}), and the previously mentioned changes in spellings of compounds (\form{lifestyle:life-style}, \form{daycare:day-care}). 

\begin{table}[!t]
    \centering
    \resizebox{\columnwidth}{!}{
    \begin{tabular}{|c | c c | }
        % \toprule
        \hline
        Neologism & \multicolumn{2}{c|}{Nearest \hist{} neighbors} \\
        % \midrule
        \hline
        email       &	telegram  &	letter  \\
        pager      &	beeper  &    	phone\\
        blogger    &	journalist & 	columnist \\
        sitcom    &	comedy  &	movie\\
        spokeswoman  &	spokesman &	director   \\
        sushi     &	caviar  &	risotto \\
        rehab       &	detoxification & aftercare \\
        \hline
        % \bottomrule
    \end{tabular}
    }
    \caption{Nearest \hist{} neighbors of projected \modern{} embeddings for a sample of emerging words. We can see that words get projected into semantically relevant neighborhoods, and nearest neighbors can even be useful for observing the evolution of a concept (e.g. \form{pager:beeper}).}
    \label{tab:nns}
\end{table}

\section{Conclusion}
We have shown that our two hypothesized factors, semantic neighborhood sparsity and its average frequency growth rate, play a role in determining in what semantic neighborhoods new words are likely to emerge. Our analyses provide more support for the latter, conforming with prior linguistic intuition of how language-external factors (which this factor implicitly represents) affect language change. We also found evidence for the former, although it was found less significant.

Our contributions are manifold. From a computational perspective, we extend prior research on meaning change to a new task of analyzing word emergence, proposing another way to obtain linguistic insights from distributional semantics.
From the point of view of linguistics, we approach an important question of whether language change is affected by not only language-external factors but language-internal factors as well. We show that internal factors---semantic sparsity, specifically---contribute to where in semantic space neologisms emerge. To the best of our knowledge, our work is the first to use word embeddings as a way of quantifying semantic sparsity. We have also been able to operationalize one kind of external factor, technological and cultural change, as something that can been measured in corpora and word embeddings, paving the way to similar work with other kinds of language-external factors in language change.

\redtext{
An admittable limitation of our analysis lies in its restricted ability to account for polysemy, which is a pervasive issue in distributional semantics studies~\cite{faruqui2016problems}. As such, semantic neologisms (existing words taking on a novel sense) were not a subject of this study, but they introduce a potential future direction.
%since their neigborhoods are polluted by co-occurrence statistics of the old sense. 
Additional properties of word's neighbors can also be correlated with word emergence, both language-internal (word abstractness or specificity) and external; these can also be promising directions for future work. Finally, our future plans include exploration of how features of semantic neighborhoods are correlated with word obsolescence (gradual decline in usage), using similar semantic observations.
}

\section*{Acknowledgments}
We thank the BergLab members for helpful discussion, and the anonymous reviewers for their valuable feedback. 
This work was supported in part by NSF grant IIS-1812327. 

\bibliography{neology_refs}
\bibliographystyle{acl_natbib}

\end{document}

%% file: glm_table.tex
\begin{table*}\centering
% \ra{1.3}
\resizebox{\textwidth}{!}{
%\scalebox{0.85}{
\begin{tabular}{@{}p{3.2cm}llllllll@{}}\toprule
\multirow{ 3}{*}{Neighborhood size} & \multicolumn{4}{c}{Stable control set} & \multicolumn{4}{c}{Relaxed control set}\\
\cmidrule(lr){2-5} \cmidrule(lr){6-9}
& \multicolumn{2}{c}{Density} & \multicolumn{2}{c}{Growth} & \multicolumn{2}{c}{Density} & \multicolumn{2}{c}{Growth}\\ 
\cmidrule(lr){2-3} \cmidrule(lr){4-5} \cmidrule(lr){6-7} \cmidrule(lr){8-9}
& $\beta_d^{(\tau)} \times 10^{4}$ & $p$-value & $\beta_r^{(\tau)} \times 10$ & $p$-value & $\beta_d^{(\tau)} \times 10^4$ & $p$-value & $\beta_r^{(\tau)}$ & $p$-value\\ 
\midrule
Large $(\tau=0.35)$ & $1.98$ & $8.25\times 10^{-5}$ & $1.84$ & $2.35\times 10^{-80} $ & $-1.07$ & $5.63\times 10^{-4}$ & $0.61$ & $2.83\times 10^{-34} $\\
%$0.375$ & $1.30$ & $8.02\times 10^{-3}$ & $1.62$ & $3.85\times 10^{-79} $ & $-1.50$ & $9.86\times 10^{-7}$ & $0.57$ & $1.86\times 10^{-37} $\\
%$0.400$ & $0.86$ & $1.23\times 10^{-1}$ & $1.47$ & $3.75\times 10^{-79} $ & $-2.03$ & $4.93\times 10^{-9}$ & $0.54$ & $2.18\times 10^{-41} $\\
%$0.425$ & $0.54$ & $4.39\times 10^{-1}$ & $1.33$ & $5.06\times 10^{-79} $ & $-2.74$ & $3.73\times 10^{-10}$ & $0.51$ & $1.66\times 10^{-45} $\\
Medium $(\tau=0.45)$ & $0.20$ & $8.29\times 10^{-1}$ & $1.16$ & $2.92\times 10^{-80} $ & $-3.67$ & $4.00\times 10^{-10}$ & $0.46$ & $6.19\times 10^{-46} $\\
%$0.475$ & $0.50$ & $6.98\times 10^{-1}$ & $1.07$ & $3.33\times 10^{-80} $ & $-4.96$ & $1.07\times 10^{-9}$ & $0.42$ & $6.08\times 10^{-47} $\\
%$0.500$ & $1.05$ & $5.57\times 10^{-1}$ & $0.91$ & $4.48\times 10^{-78} $ & $-6.46$ & $1.26\times 10^{-8}$ & $0.38$ & $3.77\times 10^{-46} $\\
%$0.525$ & $2.93$ & $2.16\times 10^{-1}$ & $0.71$ & $3.47\times 10^{-71} $ & $-7.73$ & $8.79\times 10^{-7}$ & $0.33$ & $5.51\times 10^{-43} $\\
Small $(\tau=0.55)$ & $6.90$ & $2.90\times 10^{-2}$ & $0.70$ & $1.61\times 10^{-68} $ & $-8.92$ & $4.01\times 10^{-5}$ & $0.28$ & $1.19\times 10^{-36} $\\
\bottomrule
\end{tabular}
%}
}
\caption{\label{tab:glm} Values of the GLM coefficients and their $p$-values for different neighborhood cosine similarity thresholds $\tau$. $\beta_d^{(\tau)}$ and $\beta_r^{(\tau)}$ denote the coefficients for density and average frequency growth respectively for neighborhoods defined by $\tau$. Comparing the results for the stable and relaxed control sets, we find that for the stable controls density is only significant in larger neighborhoods, but without the stability constraint both factors are significant for all neighborhood sizes.
%Weight of the growth feature is always positive, suggesting positive correlation, and weight of the density feature is negative in the relaxed setting (where it is significant), suggesting that neology is correlated with neighborhood sparsity.
} %\todo{Full table in Appendix. Bonferroni correction?} }
\end{table*}